# Syntactico-Semantic Reasoning using PCFG, MEBN & PP Attachment Ambiguity

Shrinivasan R Patnaik Patnaikuni
Department of Computer Science & Engineering.
Walchand Institute of Technology
Solapur, India
psrpatnaik@gmail.com

Dr. Sachin R Gengaje
Department of Electronics Engineering.
Walchand Institute of Technology
Solapur, India
gosachin22@gmail.com

## Abstract

Probabilistic context free grammars (PCFG) have been the core of the probabilistic reasoning based parsers for several years especially in the context of the NLP. Multi entity bayesian networks (MEBN) a First Order Logic probabilistic reasoning methodology is widely adopted and used method for uncertainty reasoning. Further upper ontology like Probabilistic Ontology Web Language (PR-OWL) built using MEBN takes care of probabilistic ontologies which model and capture the uncertainties inherent in the domain's semantic information. The paper attempts to establish a link between probabilistic reasoning in PCFG and MEBN by proposing a formal description of PCFG driven by MEBN leading to usage of PR-OWL modeled ontologies in PCFG parsers. Furthermore, the paper outlines an approach to resolve prepositional phrase (PP) attachment ambiguity using the proposed mapping between PCFG and MEBN.

**Keywords:** Syntactico-Semantic Reasoning, MEBN, PR-OWL, PCFG, PP attachment ambiguity

## 1. Introduction

This section introduces the concepts and terminologies of PCFG (Probabilistic Context Free Grammar) and MEBN (Multi Entity Bayesian Network) theory. Section 2 puts forth the proposed mapping between PCFG and MEBN theory. Section 3 outlines and discusses an approach for resolving the prepositional phrase (PP) attachment ambiguity.

A PCFG (Probabilistic Context Free Grammar) [1], probabilistically driven context free grammar is a quintuple $G_{PCFG} = (M, T, R_{PCFG}, S, P)$, where

- $M_{PCFG} = \{N^i : i = 1, . . . , n\}$ is a set of nonterminals
- $T_{PCFG} = \{w^k : k = 1, . . . , v\}$ is a set of terminals
- $R_{PCFG} = \{N^i \rightarrow \zeta^j : \zeta^j \in (M_{PCFG} \cup T_{PCFG})^*\}$ is a set of rules
- $S_{PCFG} = N^1$ is the start symbol
- $P_{PCFG}$ is a corresponding set of probabilities on rules such that

  $$\forall i \sum_j P(N^i \rightarrow \zeta^j)=1$$
- For a PCFG in chomsky normal form (CNF)
  - $R_{PCFG}=\{N^i \rightarrow N^r N^s, N^i \rightarrow w^k\}$
  - $\forall i \sum_{r,s} P(N^i \rightarrow N^r N^s) + \sum_k P(N^i \rightarrow w^k)=1$

A simple PCFG in CNF would look like

| | |
|---|---|
| S→ NP VP 1.0 | NP→ <some noun> 0.1 |
| NP→ NP PP 0.4 | NP→ <some noun> 0.04 |
| PP→ P NP 1.0 | NP→ <some noun> 0.18 |
| VP→ V NP 0.7 | NP→ <some noun> 0.1 |
| VP→ VP PP 0.3 | V→ <some verb> 1.0 |
| NP→ <some noun> 0.18 | P→ <some preposition> 1.0 |

A MEBN (Multi Entity Bayesian Network) theory [3] $T$ is a set of MFrags $\{F_1,F_2,F_3,....F_n\}$.

A MFrag $F_i$ is a quintuple $F_i = (C^i_{MEBN}, I^i_{MEBN}, R^i_{MEBN}, G^i_{MEBN}, D^i_{MEBN})$ where

- $C^i_{MEBN}$ is a finite set of values a context can take form as a value, context serves as a constraints.
- $I^i_{MEBN}$ is a set of input random variables
- $R^i_{MEBN}$ is a finite set of resident random variables
- $G^i_{MEBN}$ is a directed acyclic graph representing the dependency between input random variables and resident random variables conditional on context random variables in one to one correspondence similar to bayesian network.
- $D^i_{MEBN}$ is a set of local conditional probability distributions where each member of $R^i_{MEBN}$ has its own conditional probability distribution in set $D^i_{MEBN}$.
- Sets $C^i_{MEBN}$, $I^i_{MEBN}$, and $R^i_{MEBN}$ are pairwise disjoint.

MEBN Theory is queried using first order logic constructs , connectives, and operators. Every query on MEBN involves the construction of situation specific Bayesian network (SSBN) from the set of MFrags belonging to the concerned MTheory. MEBN theory has been widely adopted and used in various fields [2]. MEBN theory has been mapped with Relational Model of Relational Databases [11].

Given the probabilistic approach of formally defined systems like PCFG and MEBN theory for uncertainty reasoning in syntactic and semantic aspects respectively. A mapping between these two formal systems lays the foundation for syntactico-semantic reasoning.

## 2. Mapping between PCFG and MEBN:

In order to establish a connecting link between PCFG and MEBN, we need to find mapping between the members of quintuples of PCFG and MEBN. The process can be outlined in two steps.

> Step 1: Mapping of Non Terminals, Terminals in PCFG to the set of Context, Input, and Random variables in the MEBN theory.
>
> Step 2: Mapping between PCFG rule probabilities and set of local conditional probability distributions defined in MEBN.

**Step 1:**

For a mapping to exist between a PCFG and a MEBN , primarily there has to be a relation between the non terminals and terminal symbols of the entire system.

Every Non Terminal shall have an equivalent input variable in the MEBN set of MFrags such that,

$$M_{PCFG} \subset UI^{i}_{MEBN}$$

If not, shall be introduced.

Every derivation of grammar rule shall be part of ε a infinite set of entity identifiers symbols across all MFrags of the MEBN theory. ε being a infinite set of entity identifier symbols under MEBN theory,

$$\forall(N^{i} \rightarrow N^{r}N^{s}) \in R_{PCFG},\ \forall(N^{i} \rightarrow w^{k}) \in R_{PCFG},\ N^{r}N^{s} \in \varepsilon \text{ and } w^{k} \in \varepsilon$$

An additional resident random variable *hasProbability*($\theta_i,\theta_k$), where $\theta_i$ and $\theta_k$ are ordinary variables of MFrag $F_{i,}$ dependant on input random variables shall be introduced in every MFrag $F_i$ of MEBN theory.

**Step 2:**

In MEBN theory if $\{\varepsilon_1,\varepsilon_2,...,\varepsilon_n\}$, a non empty finite set of entity identifier symbols then a partial world *W* of a resident random variable *RV* is the set of all instances of the parents of random variable *RV* and the context variables of the MFrag $F_i$ that can be obtained by substituting $\varepsilon_i$ for ordinary variables $\{\theta_i\ldots\theta_n\}$ of $F_i$. A partial world state $S_W$ for partial world *W* is the set of assignments of values for each one of the random variable of the MFrag $F_i$ in the partial world.

A local distribution $\pi_{RV}$ for a resident random variable RV in MFrag $F_i$ in addition to specifying a subset of values for *RV(ε)* provides a probability distribution function $\pi_{RV(\varepsilon)}(\alpha|S_W) >= 0$ and $\sum_\alpha \pi_{RV(\varepsilon)}(\alpha|S_W)=1$, where $\alpha$ is finite subset which ranges over the set $\{\varepsilon_1,\varepsilon_2,...,\varepsilon_n\}$ *U* $\{T,F\}$. “T”, “F” denote truth values TRUE and FALSE respectively.

Combined probability distribution across PCFG and MEBN theory shall be based on conflation of probabilities from PCFG and MEBN theory.

$$P_{PCFG\text{-}MEBN}(N^i\rightarrow N^rN^s) = \&(\ P(N^i\rightarrow N^rN^s)\ ,\ \pi_{RV(\varepsilon)}(\alpha|S_W))\ where\ N^rN^s \in \alpha$$

$$P_{PCFG\text{-}MEBN}(N^i\rightarrow w^k) = \&(\ P(N^i\rightarrow w^k)\ ,\ \pi_{RV(\varepsilon)}(\alpha|S_W))\ where\ w^k \in \alpha$$

**&()** is a probability conflation function[4] which combines probabilities and normalizes with least shannon’s information loss.

For a MEBN theory *T,* a set of MFrags $\{F_1,F_2,F_3,....F_n\}$, there exists a joint unique probability distribution on the set of instances of the random variables of its MFrags that is consistent with the local probability distributions assigned within the MFrag[3].

## 3: Prepositional Phrase (PP) Ambiguity resolution using Syntactico-Semantic reasoning

Probabilistic Grammars were further augmented with Markov Random fields [7] CFGs and PCFGs focus on syntactic and structural patterns while completely ignoring semantic context during parsing.

Parsing process of natural language processing involves complex tasks which primarily include processing input sentences input and creating a representation of the sentence that particularly includes the syntactic and semantic relations. The computation tasks excel at finding syntactic relations yet are ambiguous and can't outperform the elegance displayed by the human brain in processing syntactic and semantic relations without ambiguity. On particular form of ambiguity is prepositional phrase (PP) attachment ambiguity. For example a sentence;

S:"Alex saw the girl with the telescope".

Readers typically attach the PP "with the telescope" directly to the verb phrase (VP) "saw the girl" so that the telescope is the instrument to see the verb object. The other way of interpreting the sentence attaches the PP "with the telescope" to the noun phrase (NP) "girl" so that the telescope modifies the girl with a meaning being that the girl possesses the telescope. PP attachment ambiguity is a problem for parsing. As it is syntactically ambiguous, to resolve the ambiguity properly, some form of semantics is needed. While even the entire semantics of a sentence are insufficient to solve the ambiguity.

[14] presented a survey of various literature pertaining to prepositional phrase attachment ambiguity. The survey primarily categorized the approaches to resolve PP ambiguity into categories, namely corpus based, statistical, psycholinguistic motivated and heuristic approaches. [13] is the recent attempt to solve prepositional phrase attachment using token embeddings generated from WordNet with presumption of availability of lexical ontology in natural language processing.

Based on the mapping between PCFG and MEBN defined in section 2, an approach for resolving the PP attachment ambiguity is outlined here.

Given two sentences S1:“Alex eats fish with fork” and S2:“Alex eats fish with eggs” and their possible parse trees with a sample PCFG grammar is shown in figure 3.1 and 3.2 respectively.

S→ NP VP 1.0
NP→ NP PP 0.5
PP→ P NP 1.0
VP→ V NP 0.7
VP→ VP PP 0.3
NP→ fish 0.18
NP→ eggs 0.1
NP→ fork 0.04
NP→ Alex 0.18
V→ eats 1.0
P→ with 1.0

Going on the semantic meaning of the sentences the correct prepositional phrase attachment is shown in green colored nodes in the parse trees. From the parse trees it is observed that syntactically the structure of sentences is same but semantically the meaning of sentences is different. For sentence S2 the attachment of prepositional phrase “with an egg” with noun phrase is correct syntactically and semantically. For sentence S1 the prepositional phrase “with a fork” attachment to the verb phrase is semantically correct.

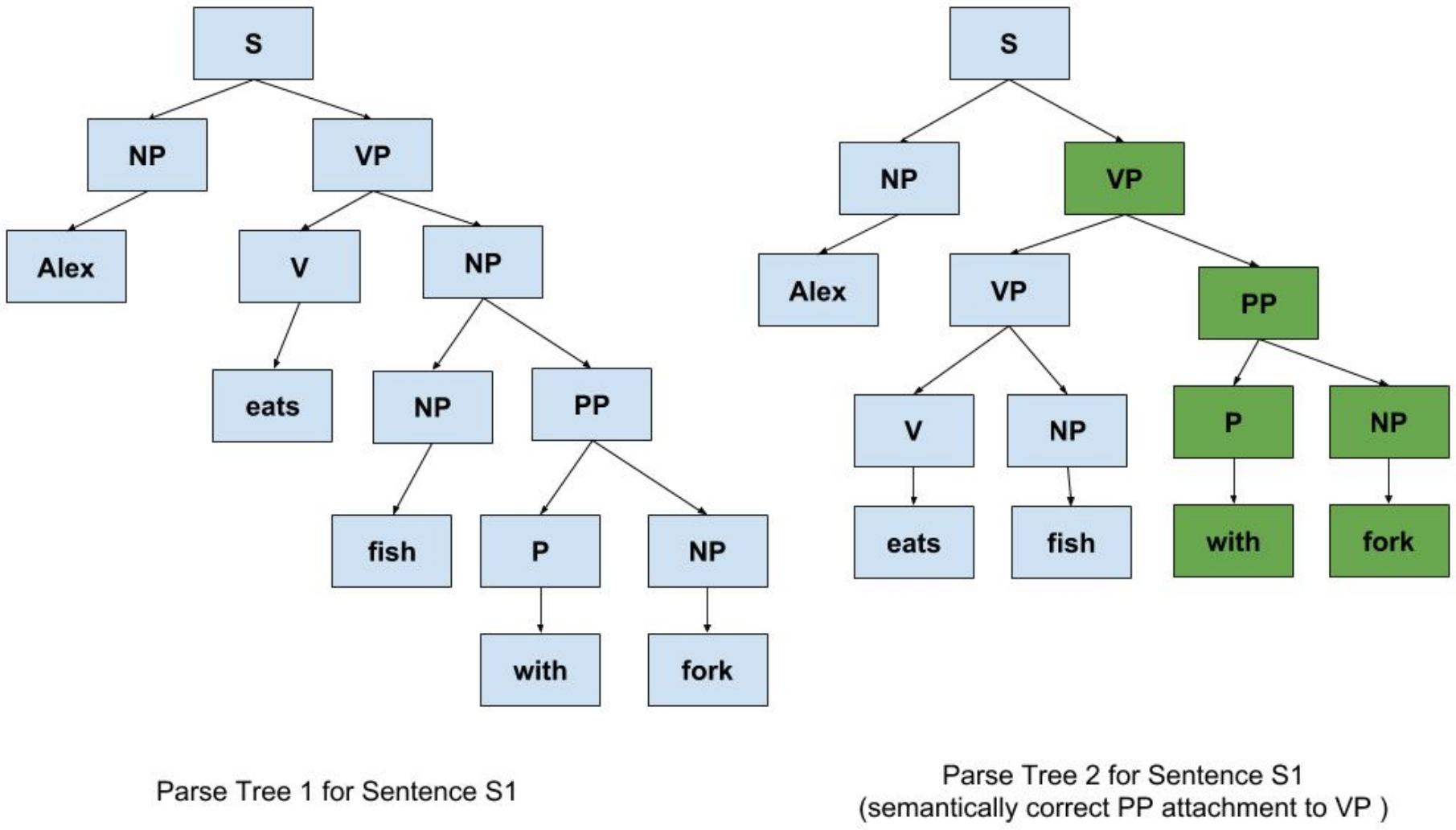

Figure 3.1: Parse trees for sentence S1

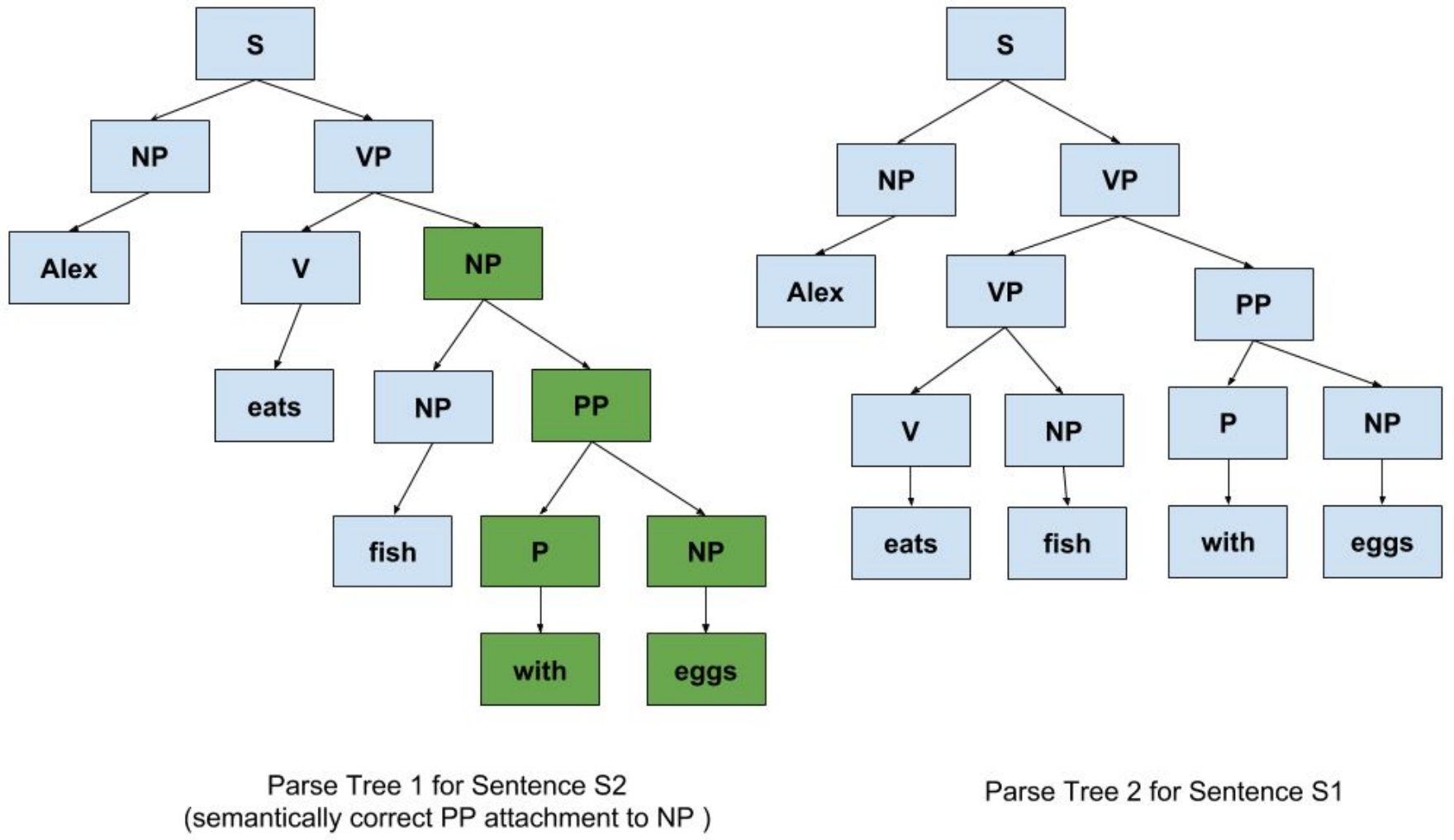

Figure 3.2: Parse trees for sentence S2

The PP attachment ambiguity can be resolved at a step of CYK algorithm where the algorithm selects a production rule with max product of probabilities.

During a step where CYK algorithm has to decide among two rules to choose from, R1, R2 with probability products $P_{PCFG}1$ and $P_{PCFG}2$ respectively for a derivation string D, a MEBN is queried on resident random variables hasProbability(D, R1), hasProbability(D, R2). The query responses produce probability values $P_{MEBN}1$, $P_{MEBN}2$ respectively. $P_{MEBN}1$, $P_{MEBN}2$ are the bayesian inference values of SSBNs generated for the queries, indicating with what probability value the string D is semantically correct derivation from production rules R1 and R2.
The MEBN queries take into account the knowledge represented using ontology for the resolving the PP attachment ambiguity.

Practically, the ontology and MEBN are designed to ascertain semantic correctness of derivation D from R1 alone is sufficient to resolve PP attachment ambiguity with MEBN query hasProbability(D, R1) producing MBNP1. From the obtained probability $P_{MEBN}1$ the CYK parser now selects production rule R1 or R2 based on condition;

*Select R1 if* $\&(P_{PCFG}1, P_{MEBN}1) > P_{PCFG}2$ *; else select R2;* where &() is function calculating conflation of probabilities.

The figure 3.3- shows the CYK parse of sentence S1 and the step where the PP attachment ambiguity resolution is attempted for sentence S1 when VP has two possible parses with probability products $P_{PCFG}1$ and $P_{PCFG}2$.

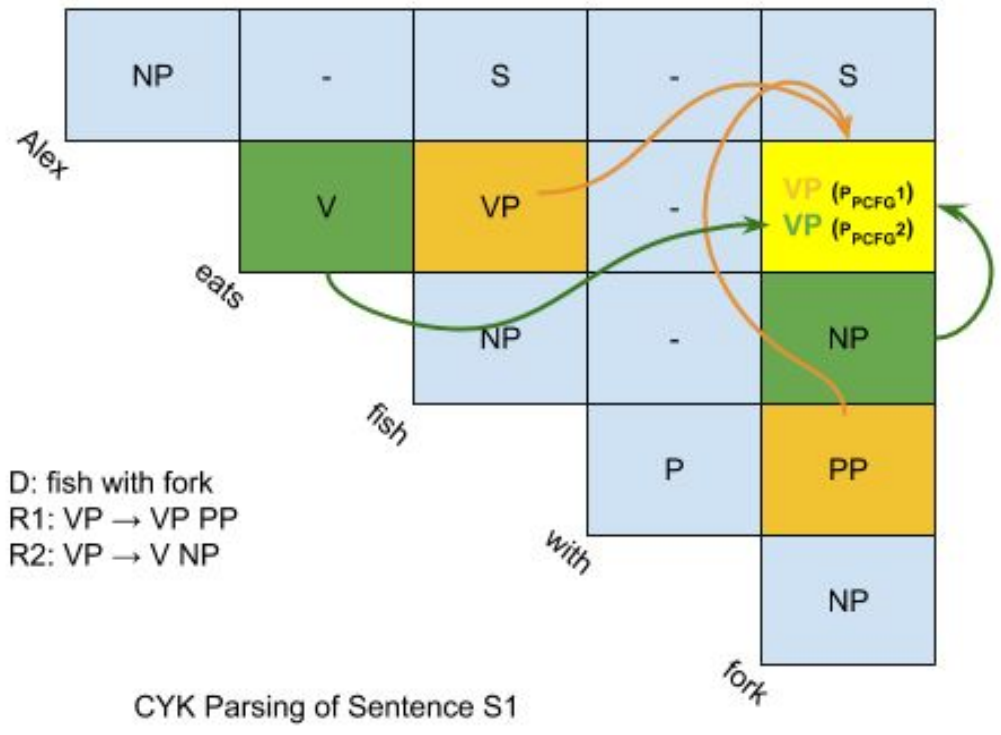

Figure 3.3: CYK Parse for sentence S1.

The MFrag of MTheory modelled on the ontology designed to semantically resolve the PP attachment to verb phrase using UnBBayes [15] is shown in figure 3.4

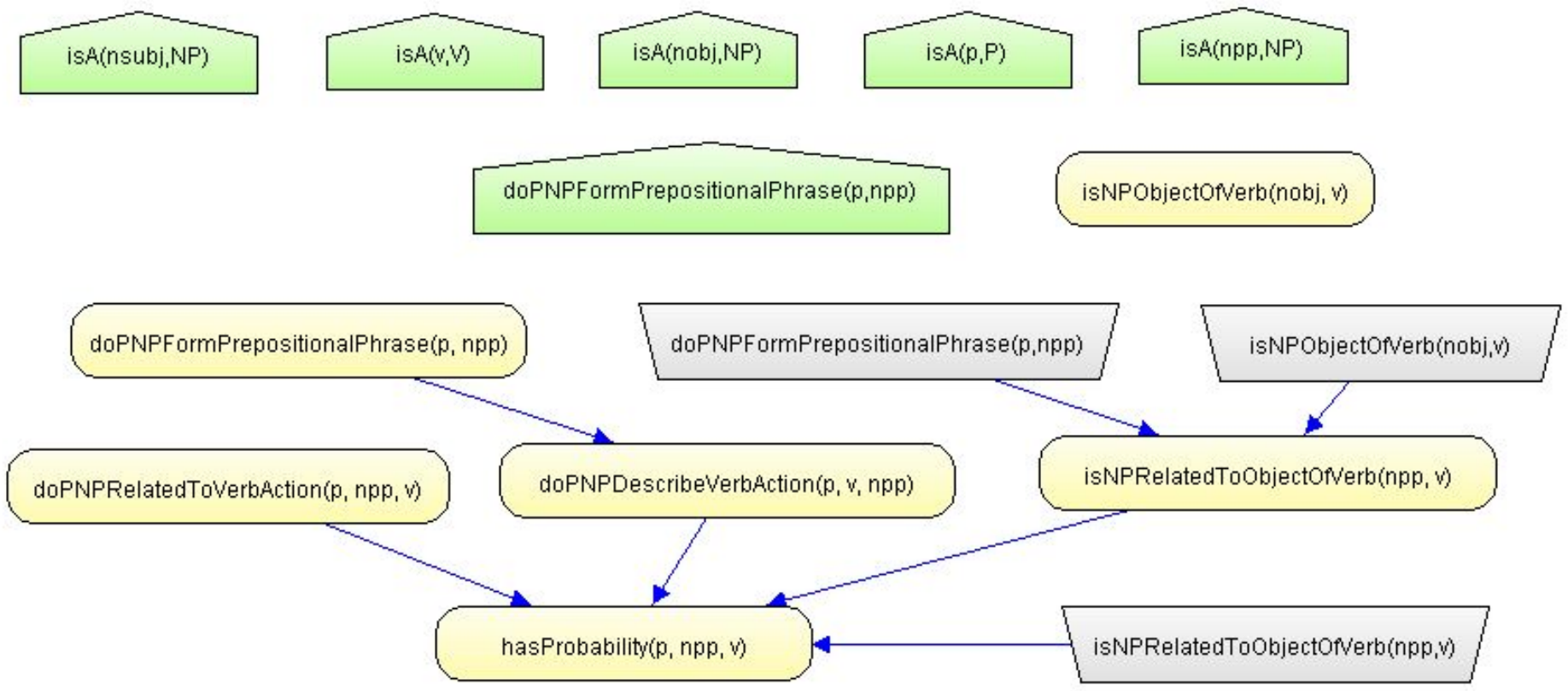

Figure 3.4: MFrag to model PP attachment.

The SSBN generated and the bayesian inference value of RV *hasProbability*($\theta_i$,$\theta_k$) is shown in figure 3.5

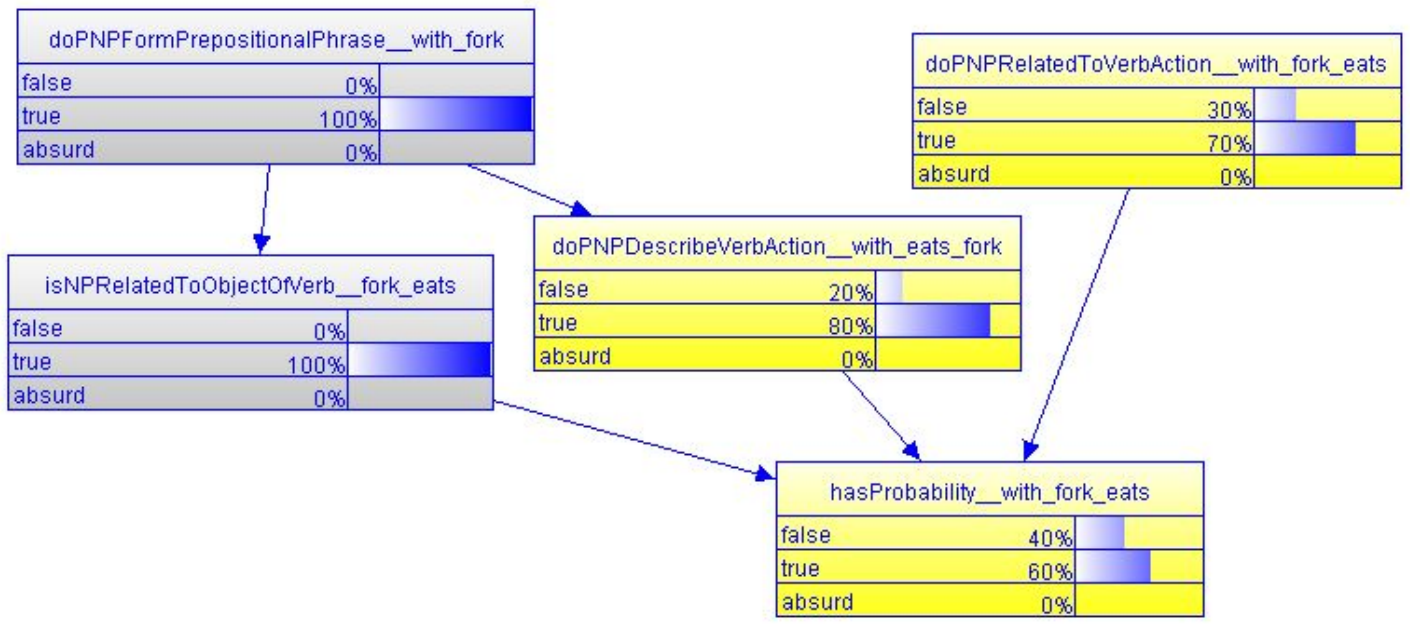

Figure 3.5: SSBN generated from MFrag to model PP attachment.

Probabilistic parsing of PCFG using a CYK parser can parse the sentences more semantically correct from the updated probabilities of grammar rules obtained by conflation of probabilities of rules and probability obtained from MEBN query on a MFrag constructed to assist the parser in resolving the ambiguity of prepositional phrase attachment to the verb phrase. The proposed method of resolving PP ambiguity has been tried on various sentences of similar nature of S1 and S2 and it has be found that more the semantic knowledge captured and encoded into MTheory's MFrags better the PP attachment ambiguity resolution.

## 4. Need for Mapping PCFG and MEBN

It is quite evident by the "Principle of Semantic Compositionality" that semantic and syntactic information goes together to give a sentence a meaning understandable by humans [8]. Though it's a different opinion that "Principle of Semantic Compositionality" is contested for it being not true all the times especially when context is introduced in the semantic meaning [9]. The work by [10] clearly points out differences in the way, how the "Principle of Semantic Compositionality" should be interpreted.

Conventionally PCFGs have been very useful in probabilistic reasoning in the context of natural language parsing and often natural languages processing tasks benefited from PCFGs. The rationale behind using probability with CFGs is to resolve a most probable parse sequence from possible set of parses available for a given string. The PCFGs are very helpful in achieving the stochasticity with which humans usually process the pattern recognition problems which includes parsing and interpretation of the meaning of sentences.

Currently several semantic information and knowledge bases are modelled as ontologies. OWL is the main language to build ontologies. The proposed integration of PCFG and MEBN theory paves a way towards usage of PR-OWL [6] with PCFG since PR-OWL is an upper ontology for specifying probabilistic ontologies expressed as MEBN Theory. Probabilistic ontologies capture the inherent uncertainties of the real world information, enabling probabilistic reasoning through MEBN theory modelling of the ontology classes, properties and relations. [12] defined and described the process of constructing probabilistic ontologies using MEBN theory, which they called PR-OWL (Probabilistic-Ontology Web Language). The approach proposed in the paper is an attempt to enable PCFGs parse sentences taking into consideration the semantic information associated with.

## 5. Summary

PCFGs have been the answer to the amphibology in natural language. MEBN theory has evolved as an extension to Bayesian networks with first order logic expressivity. MEBN theory is used in constructing probabilistic ontologies for uncertainty reasoning in knowledge bases and semantic information databases modelled as ontologies. Mapping between PCFG and MEBN would serve as a bridge linking the probabilistic reasoning in syntactic, structural oriented systems to probabilistic reasoning in semantic and probabilistic ontologies, a syntactico-semantic reasoning. Additionally, PR-OWL being the frontrunner in modelling probabilistic ontologies using MEBN's first order logic expressivity can augment with PCFGs using the mapping described in this paper. Further, the work reported in the paper needs to more meticulously address the implementation issues for the proposed work be widely adopted in real world NLP tasks.